\documentclass[a4paper,fleqn]{cas-dc}
\usepackage[numbers]{natbib}
\newcommand{\gray}[1]{{\color{gray}{#1}}}
\usepackage{algorithm}
\usepackage{algorithmicx}
\usepackage{algpseudocode}
\def\tsc#1{\csdef{#1}{\textsc{\lowercase{#1}}\xspace}}
\tsc{WGM}
\tsc{QE}

\begin{document}
\let\WriteBookmarks\relax
\def\floatpagepagefraction{1}
\def\textpagefraction{.001}

\shorttitle{Diverse and Tailored Image Generation for Zero-shot Multi-label Classification}

\shortauthors{Kaixin Zhang, Zhixiang Yuan and Tao Huang}  

\title [mode = title]{Diverse and Tailored Image Generation for Zero-shot Multi-label Classification}

\author[1]{Kaixin Zhang}[orcid=0009-0006-7666-864X]

\ead{kxzhang0618@163.com}

\credit{Writing - original draft, Writing - review \& editing, Investigation, Software, Formal analysis, Validation}

\author[1]{Zhixiang Yuan}
\cormark[1]

\ead{zxyuan@ahut.edu.cn}

\credit{Conceptualization, Supervision, Methodology, Writing - review \& editing}

\author[2]{Tao Huang}[orcid=0000-0002-4463-4078]
\ead{thua7590@uni.sydney.edu.au}
\credit{Supervision, Writing - review \& editing, Investigation}

\affiliation[1]{organization={School of Computer Science and Technology, Anhui University of Technology},
            addressline={1530 Maxiang Road}, 
            city={Maanshan},
            postcode={243032}, 
            state={Anhui},
            country={China}}

\affiliation[2]{organization={School of Computer Science, Faculty of Engineering, The University of Sydney},
            addressline={J12 Cleveland St}, 
            city={Camperdown},
            postcode={2008}, 
            state={NSW},
            country={Australia}}

\cortext[1]{Corresponding author}

\begin{abstract}
Recently, zero-shot multi-label classification has garnered considerable attention for its capacity to operate predictions on unseen labels without human annotations. Nevertheless, prevailing approaches often use seen classes as imperfect proxies for unseen ones, resulting in suboptimal performance. Drawing inspiration from the success of text-to-image generation models in producing realistic images, we propose an innovative solution: generating synthetic data to construct a training set explicitly tailored for proxyless training on unseen labels. Our approach introduces a novel image generation framework that produces multi-label synthetic images of unseen classes for classifier training. To enhance diversity in the generated images, we leverage a pre-trained large language model to generate diverse prompts. Employing a pre-trained multi-modal CLIP model as a discriminator, we assess whether the generated images accurately represent the target classes. This enables automatic filtering of inaccurately generated images, preserving classifier accuracy. To refine text prompts for more precise and effective multi-label object generation, we introduce a CLIP score-based discriminative loss to fine-tune the text encoder in the diffusion model. Additionally, to enhance visual features on the target task while maintaining the generalization of original features and mitigating catastrophic forgetting resulting from fine-tuning the entire visual encoder, we propose a feature fusion module inspired by transformer attention mechanisms. This module aids in capturing global dependencies between multiple objects more effectively. Extensive experimental results validate the effectiveness of our approach, demonstrating significant improvements over state-of-the-art methods.
\end{abstract}

\begin{keywords}
Zero-shot multi-label learning \sep Deep generative model \sep Diffusion model \sep Synthetic data
\end{keywords}

\maketitle
\section{Introduction}\label{sec1}
Image classification, a cornerstone of computer vision, witnesses substantial progress with the rise of deep learning in recent years~\cite{he2016deep, liu2021swin, jin2022regularized, jin2022deep, huang2022knowledge, huang2022lightvit, you2020greedynas}. However, these advancements predominantly focus on single-label image classification tasks, where images are assumed to contain only a singular object. In the practical domain, real-world images are more complex, often encompassing multiple objects, concepts, or scenes, thus necessitating a shift towards addressing multi-label classification (MLC) challenges.

Despite the introduction of numerous MLC methods~\cite{you2020cross, zhou2021deep, he2023open, yang2023multi} relying on fully annotated multi-label datasets, the challenge of collecting and annotating MLC data persists in practical applications. Consequently, increasing attention is being directed towards zero-shot multi-label classification (ZS-MLC), where the primary objective is to identify previously unseen classes during the inference process. 

To achieve zero-shot MLC, previous works typically involve text features to discriminate the existence of each unseen class in the image. Classical methods \cite{huynh2020shared, ben2021semantic, narayan2021discriminative} train an image feature encoder that projects the image feature into text embedding space using images and seen label texts. Therefore, in inference, the classification of unseen labels is achieved by measuring the distances between image features and text embedding of unseen labels. However, these approaches rely on single-modal information from pre-trained text embeddings such as Glove~\cite{pennington2014glove}, overlooking the valuable knowledge embedded in visual semantic image-text pairs. Recently, some works~\cite{xu2022dual, sun2022dualcoop} propose to utilize pre-trained vision-language models such as CLIP~\cite{radford2021learning} to align textual and visual spaces. Nevertheless, these methods usually fix the weights of visual encoder and text encoder in CLIP, while neglecting the domain discrepancy between CLIP training dataset and MLC dataset. Hence, regarding the fixed image and text features, it is difficult to achieve further improvements upon current methods. Moreover, all the existing methods use the seen labels as a proxy to the unseen labels, \textit{i.e.}, they train the ZS-MLC model to classify the seen labels on seen images, and assume the model can generalize to unseen labels and images. This implicit learning also leads to inferior performance compared to the conventional explicit training on fully-annotated data.

Nowadays, deep generative models~\cite{ramesh2021zero, saharia2022photorealistic, nichol2022glide, rombach2022high} are becoming increasingly powerful in generating diverse high ﬁdelity photo-realistic samples. Therefore in this paper, in light of the above issues, we get down to investigate whether the generated images can serve as effective agents to bridge the gap between zero-shot learning and fully-supervised learning and improve the ZS-MLC accuracy. Concretely, we propose a prompt-guided image generation framework based on diffusion models that can generate images containing unseen labels for the MLC dataset, then we use the synthetic data to explicitly train the classifier. Besides, to improve the efficiency and quality of generated images, we propose three improvements to the vanilla diffusion models. (1) To diversify the generated images and improve the generalization of trained MLC model, we introduce a method based on pre-trained large language model to generate diverse, detailed, and deterministic prompts, which are used to guide diffusion models to generate better multi-label images. (2) We design a discriminator relying on the pre-trained multi-modal CLIP model to recognize whether the generated images contain the target classes, and therefore we can automatically filter those wrong generated images to prevent them from affecting the accuracy. (3) To improve the generation efficiency, we introduce a CLIP score based discriminative loss to fine-tune the text encoder in diffusion model, which adapts the text prompts to be more precise and effective in generating multi-label objects in an image. Generation on the fine-tuned text encoder significantly boosts the qualified rate. These improvements in generation effectively benefit the performance and efficiency of our method.

Simultaneously, we also introduce a global feature fusion module to adapt and enhance the visual encoder in our baseline ZS-MLC method DualCoOp~\cite{sun2022dualcoop}. To alleviate the domain gap between pre-trained CLIP visual encoder and target MLC dataset, we introduce a plug-and-play global feature fusion (GFF) module into the visual encoder. GFF can acclimate the original feature to get better performance on target MLC task without losing the pre-trained general information. We achieve this by introducing a global feature branch besides the original convolution layer in visual encoder, and smoothly fine-tune the GFFs only to keep the generalization of the original feature and avoid the catastrophic forgetting issue caused by directly fine-tuning the visual encoder~\footnote{As shown in Table~\ref{tb:GFF}, fine-tuning the visual encoder decreases mAP by 6.4\% compared to not fine-tuning (baseline) on ZSL.}. Meanwhile, we build GFF with inspiration from the attention of transformer, which helps the model to capture global dependencies between multiple objects better.

To sum up, our contributions are four-fold.
\begin{enumerate}
    \item To the best of our knowledge, we are the first to introduce diffusion models into the ZS-MLC task. This approach offers a novel solution to the ZS-MLC challenge by generating training data for unseen classes and using it during the model training. 
    \item To enhance the variety of generated images, we propose a method based on a large language model to generate semantically rich prompts, which are used to guide the diffusion model to generate multi-label images. Moreover, we design a discriminator relying on the pre-trained CLIP model to filter unqualified images, which do not contain all target objects. To improve image generation efficiency, we introduce a discriminative loss utilizing CLIP score to fine-tune the text encoder in diffusion model.
    \item In order to adapt our baseline ZS-MLC method to target MLC dataset, we design a plug-and-play global feature fusion module (GFF) into pre-trained CLIP visual encoder. The domain gap between visual encoder and MLC dataset is alleviated by fine-tuning the GFFs. Besides, GFF helps visual encoder to capture global dependencies between multiple objects and achieves better performance.
    \item We conduct experiments on MS-COCO~\cite{lin2014microsoft} and NUS-WIDE~\cite{chua2009nus} and show that our method achieves significant improvements compared to previous methods. For example, on the NUS-WIDE, our method achieves 6.1\% and 3.8\% mAP improvements over the previous state-of-the-art DualCoOp~\cite{sun2022dualcoop} under ZSL and GZSL setting, respectively.
\end{enumerate}

\section{Related work}\label{sec2}
\subsection{Zero-shot multi-label classification}
Benefiting from the advances in deep learning~\cite{liu2021swin, dosovitskiy2020image, liu2022convnet}, MLC achieves remarkable success in recent year~\cite{chen2019multi, ridnik2021asymmetric, nguyen2021modular}. ASL~\cite{ridnik2021asymmetric} proposes a special loss function for MLC tasks to alleviate the imbalance between positive and negative labels distribution. ML-GCN~\cite{chen2019multi} introduces the graph convolutional network to the MLC to capture complex label dependencies. PU-MLC \cite{yuan2023positive} introduces a positive-and-unlabeled learning scheme to alleviate the disturbance of noisy labels. However, these outstanding achievements mainly focus on supervised learning, and they cannot be easily generalized to the more challenging ZS-MLC task, which aims to recognize unseen classes in the inference stage. While in the training stage, we can only utilize the given samples of seen classes to train a classification model. 

To recognize unseen classes, some pioneering works have proposed enlightening strategies in ZS-MLC. Fu et al.~\cite{fu2014transductive} propose a self-training strategy based on transfer learning, which generalizes deep regression models trained on seen classes to unseen classes. Semantic diversity learning~\cite{ben2021semantic} considers the diversity of semantic labels in the ZS-MLC task, designing a novel loss function to exploit this information to improve the model. ML-Decoder~\cite{ridnik2023ml} introduces a new classification head based on the original transformer decoder to improve the ZS-MLC results. Recently, multi-modal pre-training models have shown the powerful transfer capability in zero-shot learning. Some researches~\cite{xu2022dual, sun2022dualcoop, he2023open} attempt to apply the contrastive language-image pre-training (CLIP)~\cite{radford2021learning} model to ZS-MLC tasks, resulting in significant improvements in the field.

\subsection{Generative models for Image Recognition}
The burgeoning field of artificial intelligence generated content (AIGC) has spurred the development of text-to-image models, including DALL-E2~\cite{ramesh2021zero}, Imagen~\cite{saharia2022photorealistic}, GLIDE~\cite{nichol2022glide}, and Stable Diffusion~\cite{rombach2022high}.

Within this context, researchers have explored the application of synthetic images generated by text-to-image models in image recognition tasks. CamDiff~\cite{luo2023camdiff} leverages the latent diffusion model to generate salient objects within camouflaged scenarios, effectively augmenting the training dataset for camouflage object detection. DA-Fusion~\cite{trabucco2023effective} extends the diffusion model to novel domains by adding and fine-tuning tokens within the text encoder, thereby expanding the training images for few-shot image classification tasks. Azizi et al.~\cite{azizi2023synthetic} conducts fine-tuning of Imagen on the ImageNet~\cite{russakovsky2015imagenet} training set, resulting in synthetic images that augment the ImageNet training data. ActGen~\cite{huang2024active} improves the generation efficiency by augmenting fewer images selected with an active learning procedure, while achieving better classification performance. TTIDA~\cite{yin2023ttida} employs GPT-2~\cite{radford2019language} to generate detailed descriptions, guiding GLIDE to produce photo-realistic images with flexibility and control for data augmentation. However, extending these approaches directly to the ZS-MLC challenge is challenging due to the inherent difficulty of the native generation model in creating multi-label images (see Figure~\ref{syn_picture} (c)). To the best of our knowledge, our work marks the pioneering introduction of the diffusion model into ZS-MLC, presenting a fine-tuning generation framework tailored specifically for generating multi-label images.

\section{Preliminaries}\label{sec3}
\subsection{Zero-shot learning}
In a conventional supervised image classification task, the category set $C$ in training dataset is exactly the same as the one in test, and the trained model is only required to recognize classes within $C$.
However, in zero-shot learning (ZSL), the category set $C$ is divided into a seen category set $C_s$ and an unseen category set $C_u$. Traditional ZSL aims to train a model using training samples that contain only the seen classes in $C_s$, while predicting classes in the unseen categories $C_u$ during inference.

In contrast, generalized zero-shot learning (GZSL) is formulated to predict test samples with categories containing both trained seen classes $C_s$ and test-only unseen classes $C_u$. GZSL is more challenging to ZSL since the model is prone to overfitting on seen classes and the prediction on unseen classes would be underweighted. 

In real-world applications, ZS-MLC is more common than typical single-label ZSL because a natural image normally contains multiple objects, concepts, or scenes that should be annotated. In recent advancements on ZS-MLC task~\cite{sun2022dualcoop, he2023open, xu2022dual}, leading approaches have used the capabilities of the pre-trained CLIP model, which is acknowledged to have robust zero-shot transfer ability.

\subsection{Diffusion models}
Diffusion models (DMs) function as probabilistic generation models, learning the data distribution by iteratively denoising variables sampled from a Gaussian distribution throughout the model's training process. Specifically, in the \textit{forward process}, Gaussian noise is injected into the input image $x_0$ cumulative $T$ step to obtain $x_t$. In contrast, in the \textit{reverse process}, the noise $x_t$ is restored to the image $x_0$ by a $T$-step denoising process. 

Latent diffusion models (LDMs)~\cite{rombach2022high} propose an autoencoder to reduce the dimension of the high-dimensional image space to the low-dimensional latent space. This enhancement significantly improves the sampling efficiency of the diffusion model. Moreover, LDMs introduce the \textit{Conditioning Mechanisms} to encode specific conditional content, such as text, images, or semantic maps. These are encoded into conditional vectors by a domain-specific encoder. Those conditional vectors are incorporated into the image generation process via the cross-attention layer, guiding the generation of images that fulfill the specific conditions. In this paper, we adopt Stable Diffusion~\cite{rombach2022high}, a variant of LDMs, as the image generation model due to its enhanced stability and efficiency.

\begin{figure*}[h]%
\centering
\includegraphics[width=\textwidth]{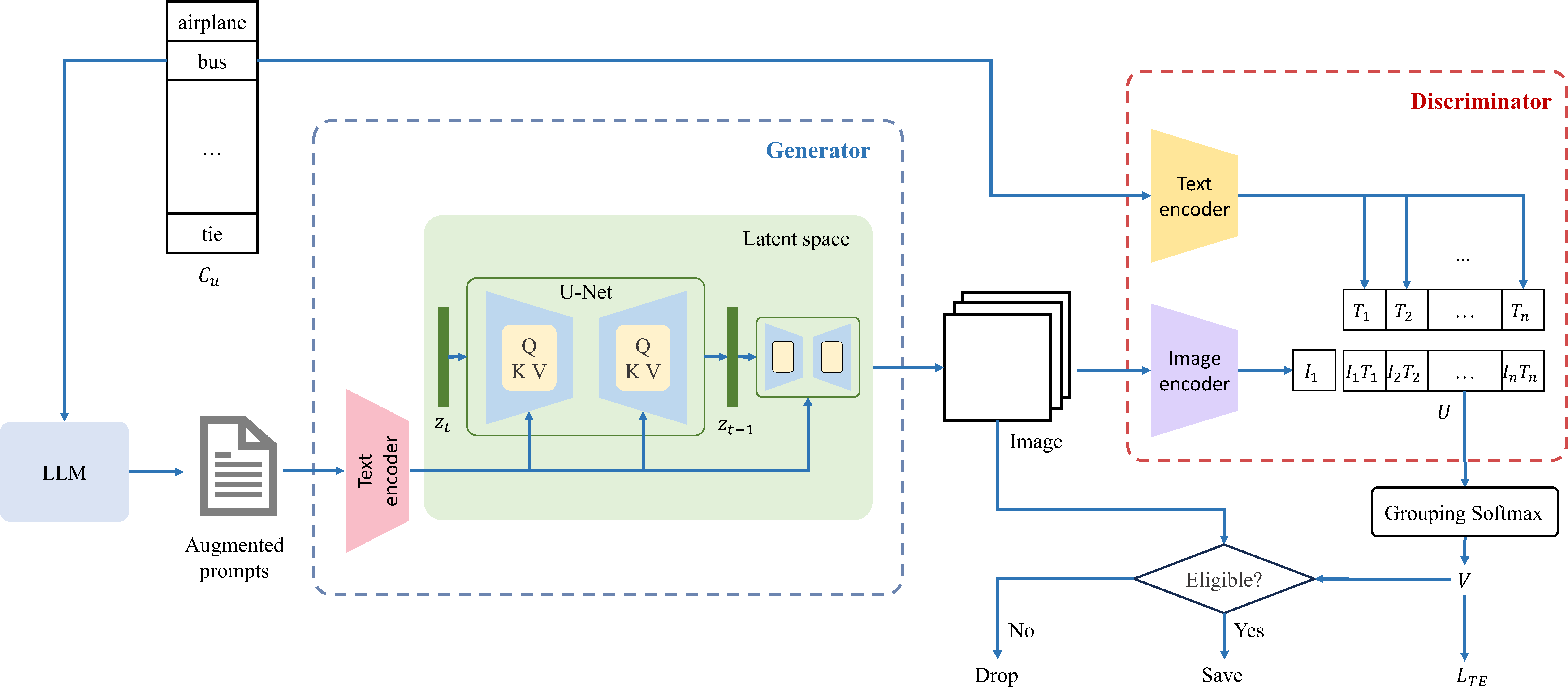}
\caption{The structure of our image generation framework}\label{farmework}
\end{figure*}
\section{The proposed approach}\label{sec4}
In the paper, we choose native Stable Diffusion~\cite{rombach2022high} as the generator and CLIP~\cite{radford2021learning} as the discriminator to structure the baseline generation framework. The steps of image generation are as follows: (1) Fill the fixed template with multiple target categories from category set $C_u$ to compose a prompt. For example, use ``cat'' and ``bus'' to form the fixed prompt ``A photo of a cat next to a bus.'' (2) The fixed prompt is modified by a LLM-based prompt enhancement method in Section~\ref{sec:4.1}, to obtain a more detailed and diverse prompt ``A cat perched on top of a bus next to a bustling city street.'' for better generating semantically rich images. (3) This augmented prompt is then leveraged as the text prompt condition to generate an image using Stable Diffusion. (4) The synthetic images are then measured by a CLIP-based synthetic image filtering method proposed in Section~\ref{sec:4.2}, and we filter those low-quality images and keep the remaining images that contain all the target classes as our training samples. The structure of our image generation framework is depicted in Figure~\ref{farmework}.

The synthetic training samples, are then extended to the original MLC dataset to get a new training dataset, which is used to replace the training dataset in our baseline ZS-MLC method in model training. Then the improved model, containing knowledge of both seen labels and unseen labels in the training set, is used to perform zero-shot predictions like the normal ZS-MLC method. The overall framework of our method is shown in Algorithm~\ref{alg:overall}.

\begin{algorithm}[t]
    \caption{Overall framework of the proposed method.}\label{alg:overall}
    \begin{algorithmic}[1]
        \Require Unseen categories $C_u$, text-to-image diffusion model $G$ with pretrained weights $\Omega$, our discriminator $R$, and MLC model $M$ with initial weights $W$;
        \State $T = \mathrm{LLM}(C_u)$; \hfill\gray{\textit{\# generate diverse prompts with in-context learning of LLM}}
        \State $\Omega^* = \arg\min_\Omega L_{ASL}(G, R, T; \Omega)$; \hfill\gray{\textit{\# fine-tune diffusion model $G$ with $L_{ASL}$, prompts $T$, and discriminator $R$}}
        \State $D_s = \mathrm{Generate}(G, R, T; \Omega^*)$; \hfill\gray{\textit{\# synthetic images with diffusion model $G$, fint-tuned weights $\Omega^*$, discriminator $R$, and prompts $T$}}
        \State $D = D_o \cup D_s$ \hfill\gray{\textit{\# combine seen training set and synthetic training set to a new dataset}}
        \State $W^* \leftarrow \mathrm{Train}(M, D; W)$; \hfill\gray{\textit{\# train the MLC model $M$ with dataset $D$}}
        \State \Return Trained MLC model with weights $W^*$.
    \end{algorithmic}
\end{algorithm}

\begin{figure*}[t]
\centering
\includegraphics[width=\textwidth]{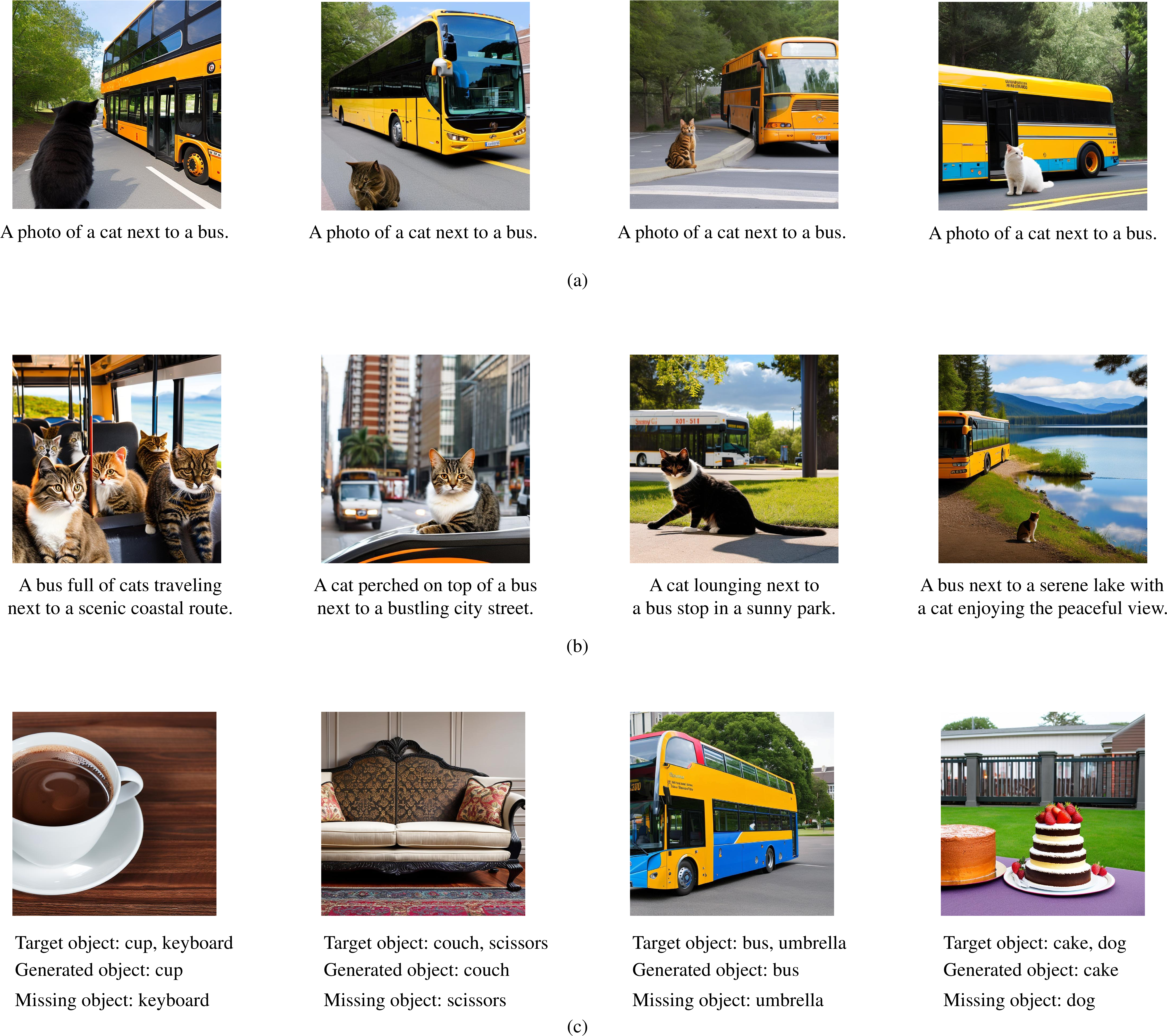}
\caption{Examples of synthetic images. \textbf{(a)} The synthetic images are generated by fixed prompts guide. \textbf{(b)} The synthetic images are generated by augmented prompts guide. \textbf{(c)} The phenomenon of missing objects persists in images generated by the diffusion model}
\label{syn_picture}
\end{figure*}

\begin{figure}[h]
\centering
\includegraphics[width=0.9\linewidth]{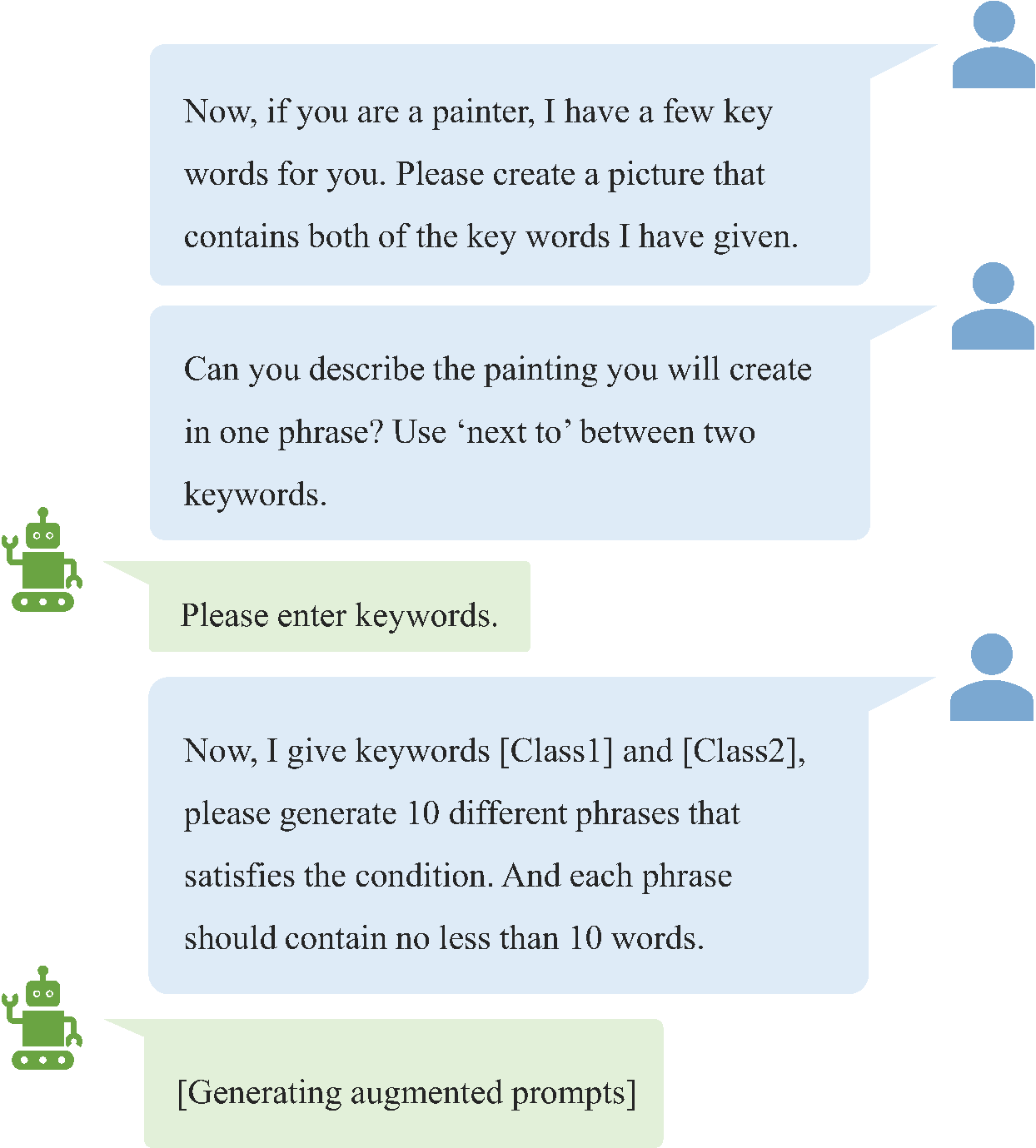}
\caption{Directs the process for the large language model to create augmented prompts}\label{augmented_prompt}
\end{figure}
\subsection{Diversifying prompts with LLMs}\label{sec:4.1}
In the domain of synthetic image generation, the prevalent use of fixed templates as prompts notably confines the variety and novelty of the generated images. This constraint manifests in the form of monotonous and uniform content, as evidenced by Figure~\ref{syn_picture} (a). To surmount this limitation, our research pioneers a novel methodology that harnesses the advanced capabilities of Llama2, a cutting-edge large language model (LLM)~\cite{touvron2023llama}, for generating semantically diverse and contextually rich language prompts. This methodology capitalizes on the in-context learning prowess of LLMs, enabling Llama2 to interpret and expand upon given textual cues in a dynamic and context-sensitive manner.

In-context learning here involves providing Llama2 with a base prompt or scenario and then guiding the model to elaborate on this scenario with additional details and variations. For example, when presented with basic object classes like ``cat'' and ``bus'', Llama2 is instructed to not just acknowledge these objects but to weave them into more complex and descriptive narratives. This process transforms a simplistic prompt such as ``A photo of a cat next to a bus.'' into an intricate and vivid description like ``A cat perched on top of a bus next to a bustling city street.'', thereby infusing the generated text with environmental context and specific object interrelations.

The technical foundation of this approach lies in the model's ability to utilize the given prompt as a contextual framework, from which it can generate a variety of semantically rich descriptions. As depicted in Figure~\ref{augmented_prompt}, we guide Llama2 using in-context instructions to generate these augmented language prompts. These prompts, once encoded by a text encoder, provide conditional vectors that direct the image generation framework towards producing a wider array of diverse and content-rich synthetic images, thus addressing the challenge of monotony and predictability in current image generation practices.

\subsection{Automatic synthetic image filtering}\label{sec:4.2}
To ensure the synthetic images used for training the multi-label classifier correctly contain the desired target objects, we employ the CLIP pre-trained model as a discriminator to filter the eligible synthetic images. With the generated image and unseen class names in $C_u$ as inputs for visual encoder and text encoder, CLIP calculates the cosine similarity $U\in\mathbb{R}^{m}$ between the image and each class name. Commonly, to choose the most adequate texts \textit{w.r.t.} to the image, $U$ is converted to a probability value $V$ via the Softmax function. For single-class image generation, we can simply check whether the probability value of the desired target class surpasses a certain threshold, to identify the correctness of generation. However, in multi-label settings, an image usually contains multiple positive labels, and calculating Softmax probabilities among them would result in mutual suppression of the positive labels. Therefore, the probabilities of positive labels are significantly smaller than the one in a single-label image, making them difficult to be prominent among all the classes.

To this end, we propose a method, called Grouping Softmax, to eliminate the interaction between positive label probabilities. First, we divide $U = [u_1,...,u_m]$ into $U_p$ for positive labels and $U_n$ for negative labels as:
\begin{align}
&U_p = [u_1^p, ...u_j^p], U_n = [u_1^n, ...u_k^n], \nonumber\\
&\text{and} \quad U = U_p \cup U_n.\label{eq1}
\end{align}
where $j$ and $k$ denote the number of positive labels and negative labels, respectively.
Each value in $U_p$ is independently reorganized with all the values in $U_n$ into the new cosine similarity sets as:
\begin{equation}
\left\{
\begin{array}{l}
     U_1 = [u_1^p, u_1^n, ..., u_k^n],\\
     ...\\
     U_j = [u_j^p, u_1^n, ..., u_k^n]. 
\end{array}\right.
\label{eq2}
\end{equation}
The Softmax function is applied independently to $U_1$ through $U_j$, ensuring that the positive label probabilities are independent of each other.
\begin{equation}
\left\{
\begin{array}{l}
     V_1 = \sigma(U_1) = [v_1^p, v_{11}^n, ..., v_{1k}^n],\\
     ...\\
     V_j = \sigma(U_j) = [v_j^p, v_{j1}^n, ..., v_{jk}^n],
\end{array}\right.
\label{eq3}
\end{equation}
where $\sigma(\cdot)$ denotes the Softmax function. Finally, we retain the positive label probabilities and average the negative label probabilities as the final output probabilities.
\begin{align}
&\left\{
\begin{array}{l}
     V_p = [v_1^p, ..., v_j^p],\\
     V_n = [\eta(v_{11}^n, ..., v_{j1}^n), ..., \eta(v_{1k}^n, ..., v_{jk}^n)],
\end{array}\right. \nonumber\\
&\text{and} \quad V = V_p \cup V_n,
\label{eq4}
\end{align}
where $\eta(\cdot)$ denotes the mean function. Set a threshold $\lambda$, and if all values in $V_p$ surpass $\lambda$ and are top-j probabilities in $V$, we consider the synthetic image as qualified and save it. The process of generating synthetic dataset is illustrated in Algorithm~\ref{alg:cap}.

\subsection{Adapting diffusion model with class discriminative loss}
By adopting augmented prompts with rich semantic information to guide image generation, the monotony of synthetic image content is alleviated significantly. However, Stable Diffusion encounters challenges in generating images containing multiple target objects, likely due to its pre-trained model lacking specialized training for such tasks. As shown in Figure~\ref{syn_picture} (c), Stable Diffusion would overlook certain target objects when generating multiple objects. In other words, not all the intended target objects are included in the synthetic image. For example, in the first image, we anticipate the presence of both the ``cup" and ``keyboard" in the generated image. Nevertheless, the ``keyboard" is omitted, leaving only the ``cup" in the generated image which obviously fall short of our expectations.

To alleviate the above issues, we fine-tune the text encoder in Stable Diffusion to suit the multi-label image generation task. Specifically, we aim to convert prompts into conditional vectors that satisfy the MLC task data domain, in order to instruct the generator to more easily generate the qualified images. During the training process, the augmented prompt is input into text encoder to get the conditional vector, which is used to guide the diffusion model to generate the image. The similarities between the generated image and each class name in $C_u$ are obtained by CLIP-based discriminator. Then, these similarities are activated by Grouping Softmax as predicted probabilities $V_p$ and $V_n$ in Equation~\ref{eq4}, which are taken as inputs to the loss function. We adopt ASL~\cite{ridnik2021asymmetric}, which is designed specifically for MLC tasks to alleviate the imbalance in the distribution of positive and negative labels, to optimize the text encoder and keep other network parameters frozen, \textit{i.e.},
\begin{align}
    \begin{split}
        \mathcal{L}_{TE}(V_p, V_n) =& \sum_{p \in V_p}(1-p)^{\gamma^+}\log{p}\\
        +&\sum_{p \in V_n}p^{\gamma^-}_c\log{(1-p_c)}
    \end{split}
    \label{eq5}
\end{align}

With the prementioned improvements in quality and efficiency of the generated images, our method can generate more diverse images such as the examples shown in Figure~\ref{syn_picture} (b).

\begin{algorithm}[t]
\caption{Image Generation for ZS-MLC.}\label{alg:cap}
\begin{algorithmic}
\Require Generated text prompts $T$ of unseen categories $C_u$, number of categories $N$ to appear in one image, fine-tuned image generator $G$ with weights $\Omega^*$, discriminator $R$, number of generated images $K$ for each category;
\end{algorithmic}

\begin{algorithmic}[1]
\State Synthetic training set $D_s = \emptyset$; 
\While{$C_u \neq \emptyset$}
    \State $C_t = \{c_i\}_{i=1}^{N}, \forall c_i \in C_u$; \hfill\gray{\textit{\# randomly sample $N$ categories for generating one image}}
    \State $t \sim T(C_t)$; \hfill\gray{\textit{\# sample a prompt corresponding to $C_t$}}
    \State $x = G(t; \Omega^*)$; \hfill\gray{\textit{\# use $t$ to generate an image}}
    \State $V_p, V_n = R(x)$; 
    \If{$V_p >= \lambda$ \textbf{and} $V_p$ in top-$N$ of $V_p \cup V_n$}
        \State $D_s = D_s \cup \{x\}$;
        \State Remove the completed category $c_i$ that has reached $K$ images in $D_s$ from $C_u$;
    \EndIf
\EndWhile
\State \Return Synthetic dataset $D_s$.
\end{algorithmic}
\end{algorithm}

\begin{figure*}[t]%
\centering
\includegraphics[width=\textwidth]{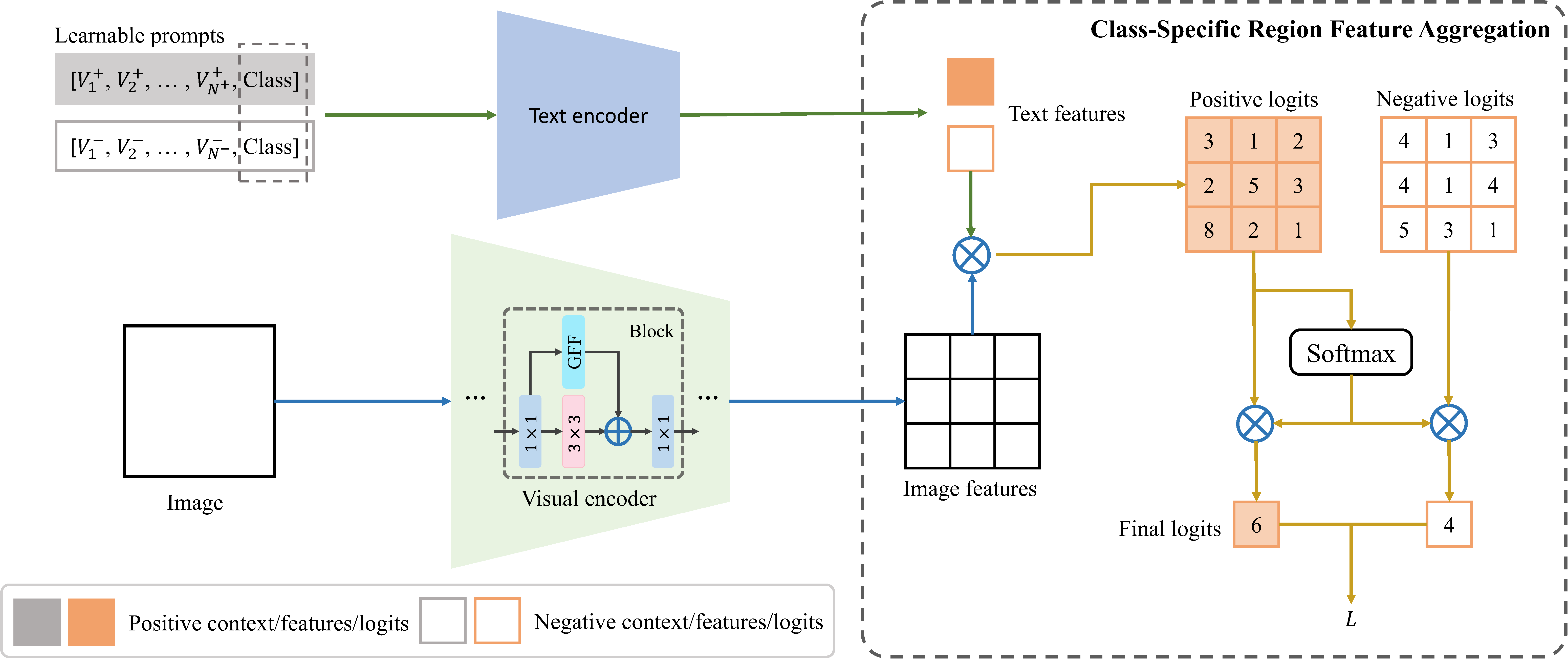}
\caption{The structure of our model. Based on DualCoOp, we introduce a global feature fusion (GFF) module that is combined with 3x3 convolutional layers in the visual encoder}\label{MLC_framework}
\end{figure*}

\begin{figure}[t]
\centering
\includegraphics[width=0.9\linewidth]{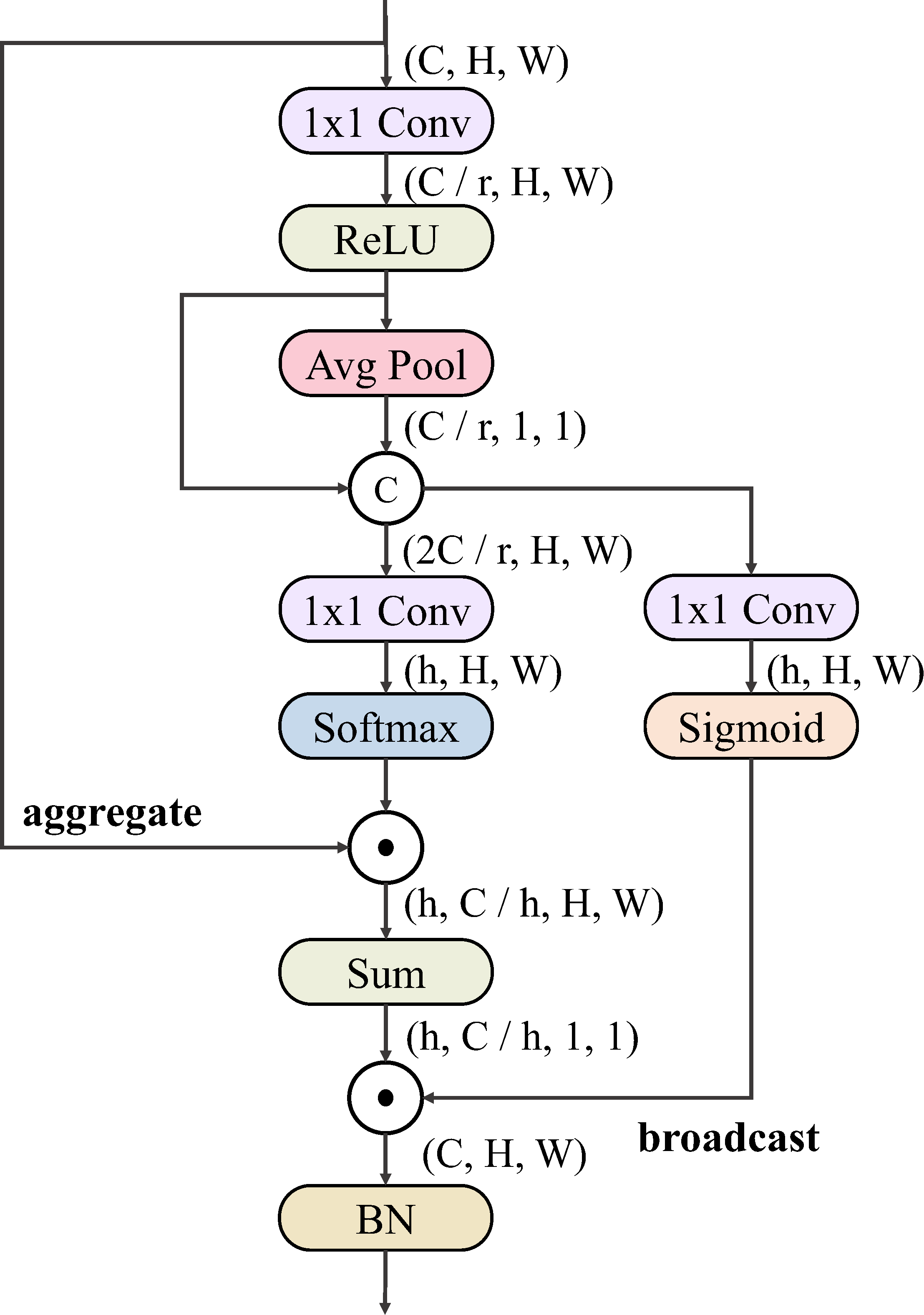}
\caption{Illustration of the global feature fusion module}\label{GFF}
\end{figure}
\subsection{Feature acclimation with global feature fusion}\label{sec:GFF}
In this paper, we choose the state-of-the-art DualCoOp~\cite{sun2022dualcoop} as the classification method for the ZS-MLC task, while our method is also capable to other zero-shot MLC models. DualCoOp introduces learnable prompts to adapt the pre-trained model to MLC datasets. Additionally, to enhance the ability to recognize multiple objects at distinct locations within the image, class-specific region feature aggregation is proposed to obtain prediction logits for each region in the image. The structure of classification method is illustrated in Figure~\ref{MLC_framework}. 

Specifically, learnable prompts are formed from a pair of contrastive prompts that provide positive and negative contextual surroundings independently for each category name. Each category is allocated a pair of contrastive prompts, called $Prompt^+$ and $Prompt^-$.

The visual feature $F_v^i$ of each region $i$ can be extracted by visual encoder $E_v$. For each region $i$ and each category $q$, the cosine similarities between visual feature $F_v^i$ and textual feature $F_t^q = E_t(Prompt)$ are expressed as:
\begin{align}
& S_{i,q}^+ = <F_v^i, (F_t^q)^+>,\nonumber\\
& S_{i,q}^- = <F_v^i, (F_t^q)^->,
\label{eq6}
\end{align}
where $<\cdot>$ denote the cosine similarity function. $S_{i,q}^+$ and $S_{i,q}^-$ are aggregated according to the magnitude of $S_{i,q}^+$ to obtain a single prediction for the entire image.
\begin{align}
& S_{q}^+ = \sum_i (\sigma(S_{i,q}^+) \cdot S_{i,q}^+),\nonumber\\
& S_{q}^- = \sum_i (\sigma(S_{i,q}^+) \cdot S_{i,q}^-).
\label{eq7}
\end{align}
Then, $S_{q}^+$ and $S_{q}^-$ are used to calculate binary classification probability $p$ as:
\begin{align}
p = \frac{e^{S_{q}^+ / \tau}}{e^{S_{q}^+ / \tau} + e^{S_{q}^- / \tau}}.
\label{eq8}
\end{align}
Finally, ASL~\cite{ridnik2021asymmetric} is used to optimize learnable prompts. Given the probabilities of positive labels $P_p$ and negative labels $P_n$ in an image, the loss is defined as
\begin{align}
\mathcal{L}_{ZS} = \mathcal{L}_{ASL}(P_p, P_n) 
\label{eq9}
\end{align}

\textbf{GFF for stable evolution of MLC model.} DualCoOp applies the classical ResNet50~\cite{he2016deep} as the visual encoder to extract image features. To mitigate the domain gap between the pre-trained visual encoder and the target MLC dataset, we endeavor to fine-tune visual encoder. However, original DualCoOp does not fine-tune visual encoder,which leads to suboptimal results. While we find that directly fine-tuning the whole visual encoder results in catastrophic forgetting. As a result, we introduce a plug-and-play module called global feature fusion (GFF) into visual encoder. Through smoothly fine-tuning the GFF, the generalization of the original features is maintained while avoiding catastrophic forgetting by directly fine-tuning the visual encoder. Meanwhile, GFF is inspired by attention of transformer, which allows the model to capture global dependencies between multiple objects in image features.

As shown in Figure~\ref{GFF}, our GFF first converts input feature into global feature that contains global information by averaging pooling. Through concat operation, the global feature and the input feature are combined into hidden feature that both contain global and local information. Then, two $1 \times 1$ convolution layers are used to produce multi-head attention in the spatial dimension and a broadcast attention, respectively, followed by Softmax and Sigmoid functions. The broadcast attention determines whether to broadcast information from a global head to each pixel. At last, the output feature is projected through a batch normalization layer, which is combined with a local feature from the $3 \times 3$ convolutional layer in the original CNNs (see Figure~\ref{MLC_framework}), resulting in the fusion of a new feature.

\section{Experiments}\label{sec5}
\subsection{Experiment settings}
\textbf{Datasets.} Consistent with previous work~\cite{huynh2020shared, ben2021semantic, sun2022dualcoop}, we perform experiments on MS-COCO~\cite{lin2014microsoft} and NUS-WIDE~\cite{chua2009nus} to validate the effectiveness of our method. Specifically, MS-COCO is partitioned into 17 unseen classes and 48 seen classes. In the NUS-WIDE dataset, 81 categories human-annotated are treated as unseen classes, while an additional 925 categories based on Flickr user tags are regarded as seen classes. Besides, it is a fine-grained dataset, which contains an average of 7 positive labels per training image, with complex correlations between many categories. For example, there exists a hierarchical relationship between ``animal'' and ``cat'', and the appearance of ``mountain'' is usually accompanied by ``rocks''. Therefore, to account for these complexities during the filtering of synthetic images, we design a specific strategy tailored for the NUS-WIDE.

\textbf{Evaluation Metrics.} Referring to previous work~\cite{huynh2020shared, ben2021semantic, sun2022dualcoop}, we report on mean average precision (mAP) and F1 score at Top-K prediction in each image on the zero-shot learning (ZSL) setting and the generalized zero-shot learning (GZSL) setting. These evaluation metrics are applied to a classifier trained on both original training images and synthetic images, to evaluate the performance enhancement in the ZS-MLC task facilitated by the synthesized data. Concretely, the F1 score is calculated by precision (P) and recall (R), and the formula is as follows:
\begin{align}
P &= \frac{TP}{TP+FP}, R = \frac{TP}{TP+FN} \nonumber\\
F1 &= \frac{2 \times P \times R}{P + R}
\label{eq:10}
\end{align} 

\textbf{Implementation Details.} The pre-trained Llama2-chat~\cite{touvron2023llama} model is used to generate augmented prompts. Concretely, we combine all unseen classes in pairs and generate 10 augmented prompts for each pair. 

We build our generation framework with Stable Diffusion V2~\cite{rombach2022high} as the generator and pre-trained CLIP~\cite{radford2021learning} with ViT-B/32 image encoder as the discriminator. With the exception of text encoder in the generator, all other parameters are frozen during training. AdamW~\cite{loshchilov2018decoupled} optimizer is adopted with LambdaLR learning rate scheduler. The hyper-parameter $j$ is assigned to 2, indicating the number of objects expected to be generated in the image. In the training stage, we randomly extract a pair of unseen classes as target objects, and randomly assign a corresponding augmented prompt as input to the text encoder. The resolution of the synthetic image is set to 560 $\times$ 560. We train our model for 4k epochs and 26k epochs on the MS-COCO and NUS-WIDE datasets, respectively, with a learning rate of 7e-8. Besides, referring to the previous work~\cite{ridnik2021asymmetric}, the hyper-parameters $\gamma^-$ and $\gamma^+$ are set to 4 and 0, respectively. 

In the inference stage, we use the same strategies as in the training stage to produce target objects and augmented prompts. These prompts are used to guide the generation of synthetic images, with 200 positive labels for each category present in the synthetic image collection. The resolution of the synthetic image is set to 768 $\times$ 768. On MS-COCO, we set the threshold $\lambda$ to 0.5 to determine whether the target object is present in the synthetic image. Especially, for the NUS-WIDE dataset, a fine-grained multi-label dataset, we set the threshold $\lambda$ to 0.1. If all values in the $V_p$ exist in the Top-k (k=7, an average of 7 positive labels in each image on NUS-WIDE) probabilities of the $V_p \cup V_n$, we treat the synthetic image as qualified and save it. Additionally, for other categories in the Top-k probabilities, if their values exceed $\lambda$, we also treat them as positive labels in the synthetic image. 

In this paper, we choose the advanced DualCoOp~\cite{sun2022dualcoop} as the training method of the multi-label classifier. Following DualCoOp, we utilize the CLIP pre-trained ResNet-50~\cite{he2016deep} as the backbone, with an input resolution of 224. We train it for 50 epochs with a batch size of 32 / 96 for MS-COCO / NUS-WIDE, respectively, setting the learning rate to 0.002 / 0.0005.

\subsection{Experiment results}
\begin{table*}[h]
\caption{Zero-Shot Multi-label Classification on NUS-WIDE}\label{tb:NUS-WIDE}
\begin{tabular*}{\textwidth}{@{\extracolsep\fill}lccccccc}
\toprule%
& \multicolumn{3}{@{}c@{}}{Top-3} & \multicolumn{3}{@{}c@{}}{Top-5} \\\cmidrule{2-4}\cmidrule{5-7}%
Method & P & R & F1 & P & R & F1 & mAP\\
\midrule
ZSL\\
\midrule
CONSE & 17.5 & 28.0 & 21.6 & 13.9 & 37.0 & 20.2 & 9.4\\
LabelEM & 15.6 & 25.0 & 19.2 & 13.4 & 35.7 & 19.5 & 7.1\\
Fast0Tag & 22.6 & 36.2 & 27.8 & 18.2 & 48.4 & 26.4 & 15.1\\
OAL & 20.9 & 33.5 & 25.8 & 16.2 & 43.2 & 23.6 & 10.4\\
LESA(M=10) & 25.7 & 41.1 & 31.6 & 19.7 & 52.5 & 28.7 & 19.4\\
BiAM & – & – & 33.1 & – & – & 30.7 & 26.3\\
SDL(M=7) & 24.2 & 41.3 & 30.5 & 18.8 & 53.4 & 27.8 & 25.9\\
DualCoOp  & 37.3 & 46.2 & 41.3 & 28.7 & 59.3 & 38.7 & 43.6\\
Ours  & \textbf{40.2} & \textbf{49.8}  & \textbf{44.5}  & \textbf{30.2} & \textbf{62.4} & \textbf{40.7} & \textbf{49.7}\\
\midrule
GZSL\\
\midrule
CONSE & 11.5 & 5.1 & 7.0 & 9.6 & 7.1 & 8.1 & 2.1\\
LabelEM & 15.5 & 6.8 & 9.5 & 13.4 & 9.8 & 11.3 & 2.2\\
Fast0Tag & 18.8 & 8.3 & 11.5 & 15.9 & 11.7 & 13.5 & 3.7\\
OAL& 17.9 & 7.9 & 10.9 & 15.6 & 11.5 & 13.2 & 3.7\\
LESA(M=10) & 23.6 & 10.4 & 14.4 & 19.8 & 14.6 & 16.8 & 5.6\\
BiAM & – & – & 16.1 & – & – & 19.0 & 9.3\\
SDL(M=7) & 27.7 & 13.9 & 18.5 & 23.0 & 19.3 & 21.0 & 12.1\\
DualCoOp  & 31.9 & 13.9 & 19.4 & 26.2 & 19.1 & 22.1 & 12.0\\
Ours  & \textbf{37.1} & \textbf{16.2}  & \textbf{22.6}  & \textbf{30.8} & \textbf{22.5} & \textbf{26.0} & \textbf{15.8}\\
\bottomrule
\end{tabular*}
\footnotetext{}
\end{table*}
\textbf{Results on NUS-WIDE.} To validate the effectiveness of our approach, we conducted comparisons with current published state-of-the-art methods, including CONSE~\cite{norouzi2014zero}, LabelEM~\cite{akata2015label}, Fast0Tag~\cite{zhang2016fast}, OAL~\cite{kim2018bilinear}, LESA~\cite{huynh2020shared}, BiAM~\cite{narayan2021discriminative}, SDL~\cite{ben2021semantic}, Deep0Tag~\cite{rahman2019deep}, and DualCoOp~\cite{sun2022dualcoop}. As the experimental results are shown in Table~\ref{tb:NUS-WIDE}, our method outperforms previous approaches across all evaluation metrics on NUS-WIDE. Compared to the state-of-the-art DualCoOp, our approach demonstrates a substantial improvement in ZSL, with the mAP increasing from 43.6\% to 49.7\%, representing a significant increase of 6.1\%. Moreover, in the more challenging GZSL, our method showcases a notable improvement in mAP, surpassing DualCoOp by 3.8\%. Moreover, our approach demonstrates superior performance with F1 scores that significantly outperform the suboptimal results in both the ZSL and GZSL settings.

\begin{table*}[h]
\caption{Zero-Shot Multi-label Classification on MS-COCO. * indicates the result we obtained by reproducing the experiment}\label{tb:MS-COCO}
\begin{tabular*}{\textwidth}{@{\extracolsep\fill}lcccccccc}
\toprule%
& \multicolumn{4}{@{}c@{}}{ZSL} & \multicolumn{4}{@{}c@{}}{GZSL} \\\cmidrule{2-5}\cmidrule{6-9}%
Method & P & R & F1 & mAP & P & R & F1 & mAP\\
\midrule
CONSE & 11.4 & 28.3 & 16.2 & - & 23.8 & 28.8 & 26.1 & -\\
Fast0Tag & 24.7 & 61.4 & 25.3 & - & 38.5 & 46.5 & 42.1 & -\\
Deep0Tag & 26.5 & 65.9 & 37.8 & - & 43.2 & 52.2 & 47.3 & -\\
SDL(M=2) & 26.3 & 65.3 & 37.5 & - & 59.0 & 60.8 & 59.9 & -\\
DualCoOp & 35.3 & 87.6 & 50.3 & 78.1$^*$ & 58.4 & 68.1 & 62.9 & 69.3$^*$\\
Ours  & \textbf{36.0} & \textbf{89.4} & \textbf{51.3}  & \textbf{83.8} & \textbf{59.2} & \textbf{69.1} & \textbf{63.8} & \textbf{72.7}\\
\bottomrule
\end{tabular*}
\footnotetext{}
\end{table*}
\textbf{Results on MS-COCO.} In Table~\ref{tb:MS-COCO}, a comprehensive comparison between our method and state-of-the-art approaches is presented on MS-COCO. In contrast to the suboptimal DualCoOp, our method also achieves significant advantages across all evaluation metrics. Especially in mAP, our method exhibits improvements of 5.7\% in ZSL and 3.4\% in GZSL. Similarly, in F1 scores, our approach outperformed others, achieving 51.3\% in ZSL and 63.8\% in GZSL, demonstrating its superior performance.

\begin{table*}[h]
\caption{Analysis on fine-tuning visual
encoder on MS-COCO}\label{tb:GFF}
\begin{tabular*}{\linewidth}{@{\extracolsep\fill}lcccc}
\toprule
& \multicolumn{2}{@{}c@{}}{ZSL} & \multicolumn{2}{@{}c@{}}{GZSL} \\\cmidrule{2-3}\cmidrule{4-5}%
Method & F1 & mAP & F1 & mAP\\
\midrule
\textit{Baseline}\\
\midrule
Fine-tune learnable prompts Only & 50.6 & 82.9 & 63.2 & 71.8\\
\midrule
\textit{Directly fine-tune visual encoder}\\
\midrule
Synchronously fine-tune learnable prompts and visual encoder & 7.6 & 9.2 & 4.0 & 5.0\\
Fine-tune visual encoder, then learnable prompts & 14.0 & 7.5 & 3.8 & 3.9\\
Fine-tune learnable prompts, then visual encoder & 48.6 & 76.5 & 61.5 & 69.0\\
\midrule
\textit{Fine-tune visual encoder through GFF}\\
\midrule
Fine-tune learnable prompts, then GFF & 49.9 & 83.5 & 63.6 & 72.2\\
Ours (synchronously fine-tune learnable prompts and GFF) & \textbf{51.3} & \textbf{83.8} & \textbf{63.8}  & \textbf{72.7}\\
\bottomrule
\end{tabular*}
\footnotetext{}
\end{table*}
\subsection{Analysis on fine-tuning visual encoder}

In this paper, we assert that directly fine-tuning the whole visual encoder will cause a catastrophic forgetting issue, which is well-studied in transfer learning. Therefore, we introduce a global feature fusion (GFF) module to smoothly evolve the visual feature. Now we give detailed empirical analyses to demonstrate our idea. We explore several fine-tuning options of the visual encoder as follows.

\begin{itemize}
    \item \textbf{Directly fine-tuning visual encoder with prompts.} In the original DualCoOp method, only prompt tokens are learnable during training. To adapt visual encoder with target MLC training set, the most straight-forward way is directly fine-tuning the whole visual encoder along with the prompt tokens in training. However, as shown in Table~\ref{tb:GFF}, synchronously fine-tuning learnable prompts and visual encoder leads to a severe collapse in precision. This phenomenon may be attributed to the co-optimization of learnable prompts and the visual encoder, causing the model to converge to local minima that over-fits the small-scale training samples. 
    
    \item \textbf{Optimizing learnable prompts and the visual encoder sequentially.} To avoid the mutual-influence between visual encoder and prompt tokens, we attempt to decouple the training of them, which leads to two orders of two-stage training manner: (1) first train visual encoder and fix prompts, then fix visual encoder and train prompts; (2) first train prompts, then train visual encoder. However, the (1) also performs poorly due to the absence of useful semantic information in the initial learnable prompts, and directly fine-tuning the visual encoder leads to meaningless fitting of the visual features and random text features. In contrast, (2) performs much more normal, but still largely behind the baseline method without fine-tuning on visual encoder.

    \item \textbf{Fine-tuning GFFs with prompts.} To prevent the adverse impact of directly fine-tuning visual encoder, we adapt visual encoder to MLC datasets by optimizing GFFs, thereby preserving the generalization capabilities of visual encoder. Meanwhile, GFF aids visual encoder in capturing global features and injecting global dependencies into the model's output features. As shown in Table~\ref{tb:GFF}, we optimize learnable prompts and GFFs synchronously during model training and achieve a significant improvement over baseline and directly fine-tune visual encoder. This demonstrates that GFFs avoid catastrophic forgetting resulting from the direct fine-tuning of visual encoder, while the global dependencies extracted by GFFs contribute to improved performance in classification tasks. 

    \item \textbf{First prompts, then GFFs.} In the experiment involving fine-tuning visual encoder, optimizing learnable prompts first and then fine-tuning visual encoder results better performance. Therefore, we conduct a similar experiment to assess the impact of the two-stage training strategy on GFFs. However, this two-stage training strategy does not yield significant improvements in classification accuracy. Meanwhile, it increases the time cost of training model, as a classification model needs to undergo two complete training processes.
\end{itemize}

Consequently, in this paper, we employ a simple and efficient strategy of optimizing synchronously learnable prompts and GFFs to train the classification model.

\subsection{More ablation study}
\begin{figure*}[h]
\centering
\includegraphics[width=\textwidth]{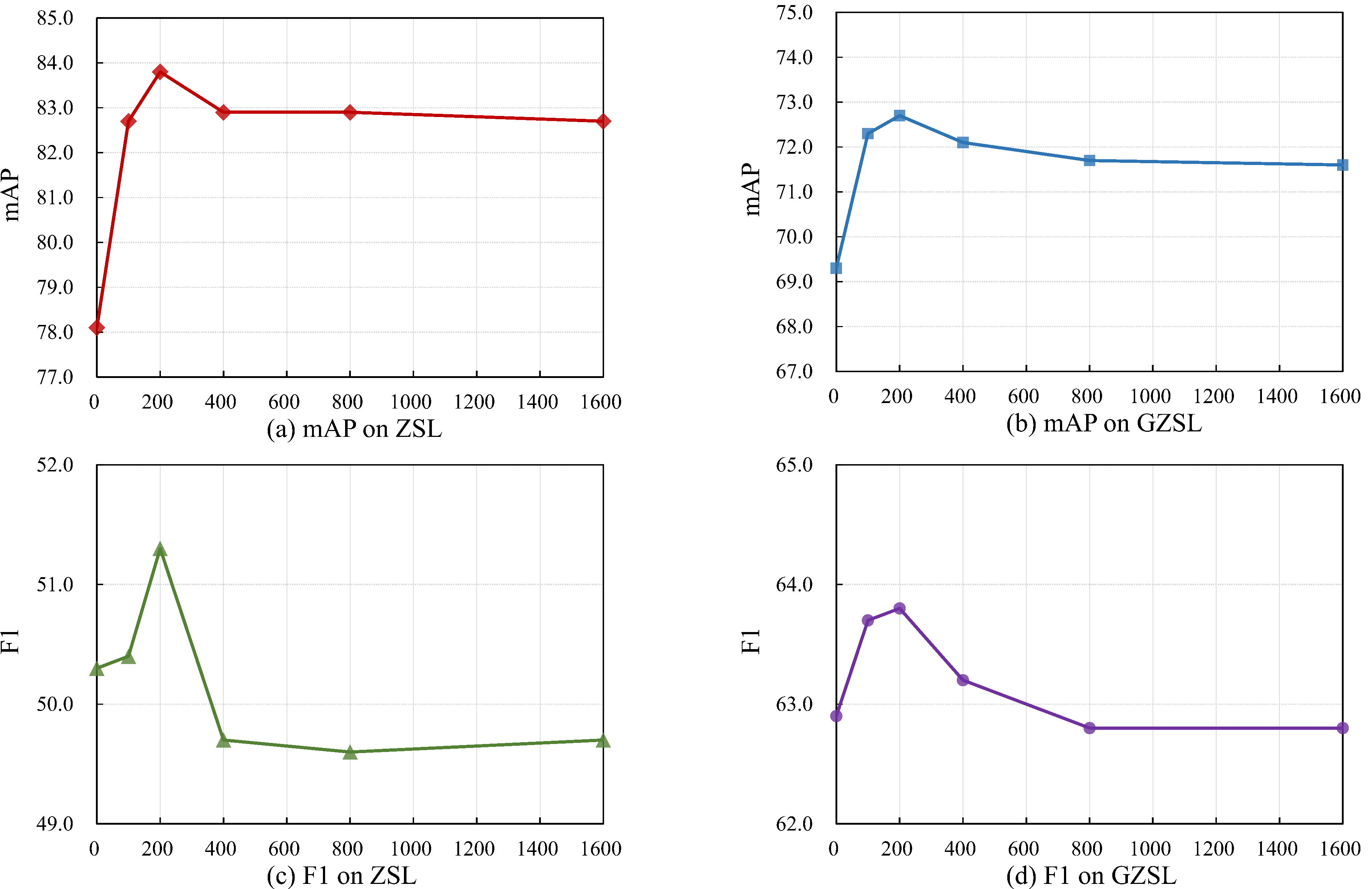}
\caption{Ablation study on the number of synthetic images on MS-COCO}\label{category_number}
\end{figure*}
\textbf{Ablation study on the number of synthetic images.} To analyze the impact of the number of synthetic images on ZS-MLC, we conducted a series of experiments on the MS-COCO dataset. Specifically, we generate a series of image collections, which are used separately for classifier training, with 100/200/400/800/1600 positive labels per unseen class. As Figure~\ref{category_number} shows, when the number of per unseen class is 200, the classification method makes the most gains in mAP and F1 scores. An excessive number of synthetic images can diminish the classifier's enhancement on the mAP and directly reduce the F1 score, particularly when the number of positive labels per unseen class equals or exceeds 400. The reason for this phenomenon may be attributed to disparities between the synthetic images and the real images in the test set. Superabundant synthetic images lead the classifier to overfit the virtual category representation, thereby impeding the classifier's ability to recognize real data.

\begin{figure*}[h]
\centering
\includegraphics[width=0.9\textwidth]{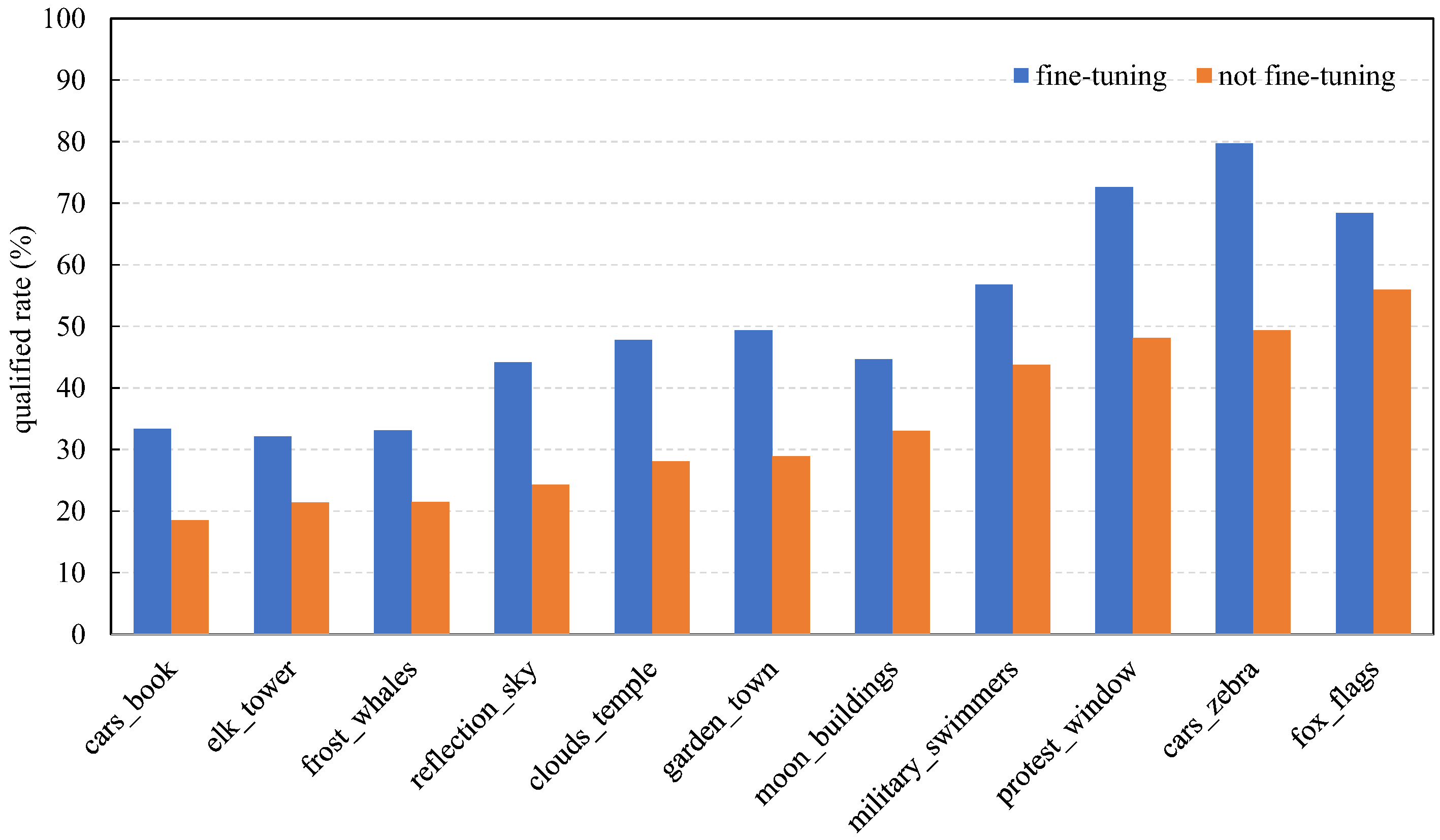}
\caption{The qualified rates of partial category pairs on NUS-WIDE}\label{qualified_rate}
\end{figure*}
\textbf{Ablation study on fine-tuning text encoder.} Our aim in fine-tuning the text encoder is to enhance the qualified rate of synthetic images, especially for challenging category pairs. The experiment is conducted on certain category pairs in which objects are easily overlooked in the generated images. Specifically, we generate 1k images for these challenging category pairs using our generation framework with and without the fine-tuned text encoder, respectively, and calculate their qualified rates. Figure~\ref{qualified_rate} illustrates a significant increase in the qualified rate for those challenging category pairs after fine-tuning the text encoder. For instance, the category pair ``cars-zebra" witnesses an improvement of over 30\% in the qualified rate, and ``protest-window" achieves a qualified rate increase of more than 20\%. Besides, the enhancement of the qualified rate is not less than 10\% on the remaining category pairs.

\begin{table}[h]
\caption{Comparisons of different prompts to guide image generation on MS-COCO}\label{tb:augmented_prompts}
\begin{tabular*}{\linewidth}{@{\extracolsep\fill}lcccc}
\toprule
& \multicolumn{2}{@{}c@{}}{ZSL} & \multicolumn{2}{@{}c@{}}{GZSL} \\\cmidrule{2-3}\cmidrule{4-5}%
Method & F1 & mAP & F1 & mAP\\
\midrule
Fixed prompts & 49.8 & 82.0 & 63.0 & 72.0\\
Augmented prompts & \textbf{51.3} & \textbf{83.8} & \textbf{63.8}  & \textbf{72.7}\\
\bottomrule
\end{tabular*}
\footnotetext{}
\end{table}
\textbf{Ablation study on the augmented prompts.} As shown in Figure~\ref{syn_picture} (b), the most intuitive effect of the augmented prompts is to significantly improve the semantic richness of the synthetic image. Absolutely, the ultimate purpose of image generation is to augment the training process of the classification method. Therefore, we conduct experiments to verify the effect of synthetic images, generated with the guidance of augmented prompts, on the accuracy of classification method. As illustrated in Table~\ref{tb:augmented_prompts}, compared with fixed prompts, images generated using augmented prompts substantially enhance the performance of the classification method. The F1 scores and mAP improved by 1.5 and 1.8\% respectively in ZSL, indicating that diverse synthetic images enhance the generalization ability of the classification model.

\begin{table}[h]
\caption{Ablation on multi-label classification method on MS-COCO}\label{tb:ASL}
\begin{tabular*}{\linewidth}{@{\extracolsep\fill}lcccc}
\toprule
& \multicolumn{2}{@{}c@{}}{ZSL} & \multicolumn{2}{@{}c@{}}{GZSL} \\\cmidrule{2-3}\cmidrule{4-5}%
Method & F1 & mAP & F1 & mAP\\
\midrule
\textit{Without synthetic images}\\
\midrule
ASL & 10.2 & 9.1 & 51.2 & 52.9\\
\midrule
\textit{With synthetic images}\\
\midrule
ASL & 43.0 & 62.5 & 55.3 & 58.0\\
DualCoOp & \textbf{50.6} & \textbf{82.9} & \textbf{63.2} & \textbf{71.8}\\
\bottomrule
\end{tabular*}
\footnotetext{}
\end{table}
\textbf{Comparison of using synthetic images on MLC and ZS-MLC frameworks.} In this paper, we generate training samples for the unseen classes and extend them into the seen training set to get the training dataset. Our model framework uses the model framework for zero-shot MLC prediction, while an interesting question is that, \textit{if we use the new training dataset to train a classical MLC model with fully-supervised setting, can it perform on pair with training MLC with all real images?}  Here we contemplate employing the new training set for supervised learning methods. Consequently, we select ASL~\cite{ridnik2021asymmetric}, a classic MLC method, as the supervised learning method for our experiments. ASL utilizes ResNet50 as the backbone and introduces a specialized loss function for MLC tasks to mitigate the imbalance in positive and negative label distribution. As demonstrated in Table~\ref{tb:ASL}, ASL performed less effectively than DualCoOp across all evaluation metrics. While both ASL and DualCoOp utilize the same pre-trained ResNet50 as the visual encoder, DualCoOp incorporates a pre-trained text encoder, introducing additional text features during model training to enhance classification. Therefore, we apply DualCoOp, a pathbreaking ZS-MLC method, as our baseline classification method.
    
Additionally, we observe that ASL achieves disastrous performance when training the model with the original zero-shot dataset. Particularly in the ZSL task, F1 scores and mAP are only 10.2 and 9.1\%, respectively. On the contrary, when trained with the new training dataset, ASL demonstrated noteworthy improvements in F1 scores and mAP scores of 32.8 and 53.4\% on ZSL, respectively. This implies that synthetic data is effective in enhancing the model's ability to recognize unseen categories.

\begin{table}[h]
\caption{Ablation on hyper-parameters $\lambda$ on MS-COCO}\label{tb:lambda}
\begin{tabular*}{\linewidth}{@{\extracolsep\fill}lcccc}
\toprule
& \multicolumn{2}{@{}c@{}}{ZSL} & \multicolumn{2}{@{}c@{}}{GZSL} \\\cmidrule{2-3}\cmidrule{4-5}%
Method & F1 & mAP & F1 & mAP\\
\midrule
$\lambda$ = 0.3 & 50.8 & 82.7 & 63.4 & 72.3\\
$\lambda$ = 0.7 & 51.0 & 82.7 & 63.6 & 72.1\\
Ours ($\lambda$ = 0.5) & \textbf{51.3} & \textbf{83.8} & \textbf{63.8} & \textbf{72.7}\\
\bottomrule
\end{tabular*}
\footnotetext{}
\end{table}

\begin{figure*}[h]
\centering
\includegraphics[width=\textwidth]{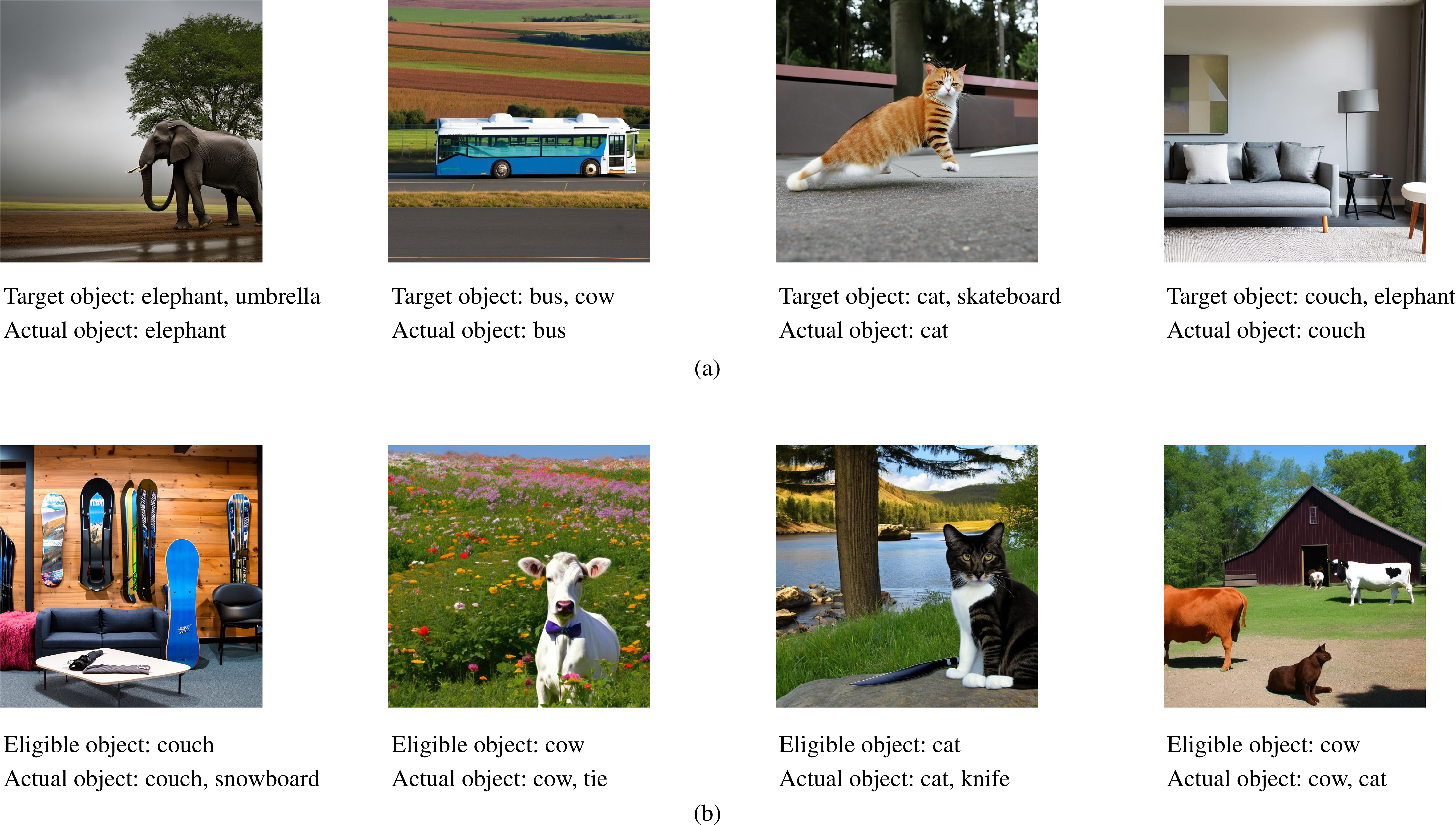}
\caption{Synthetic images misjudged by the discriminator. \textbf{(a)} A lower threshold, with $j$ set to 0.3, causes the discriminator to judge some unqualified images, which lack the target object, as qualified. \textbf{(b)} A higher threshold, like setting $j$ to 0.7, leads the discriminator to perceive certain objects in the image as non-existent, resulting in categorizing the qualified image as unqualified}
\label{threshold}
\end{figure*}
\textbf{Ablation study on hyper-parameters.} We implement experiments on MS-COCO to analyze the effects of hyper-parameters, specifically $\lambda$ and $j$. Table~\ref{tb:lambda} presents the results in mAP and F1 scores using different $\lambda$. Our setting leads to significantly higher F1 scores and mAP compared to setting $\lambda$ to 0.3. A lower filtering threshold, such as assigning $\lambda$ to 0.3, allows unqualified synthetic images to pass the discriminator (see Figure~\ref{threshold} (a)). As a result, it has a detrimental effect on the training of the classification method, which treats these unqualified images as training data. On the contrary, setting the $\lambda$ to 0.7 (a higher value) results in a more rigorous filtering process by the discriminator for synthetic images. However, it would rule some truly qualified synthetic images as unqualified (see Figure~\ref{threshold} (b)), thus reducing the efficiency of our generation framework. Meanwhile, setting $\lambda$ to 0.7 does not result in higher F1 scores and mAP compared to our setting. Therefore, we set the value of $\lambda$ to 0.5, striking a balance between rigorous filtering and preserving the efficiency of the generation framework.

\begin{table}[h]
\caption{Ablation on hyper-parameters $j$ on MS-COCO}\label{tb:j}
\begin{tabular*}{\linewidth}{@{\extracolsep\fill}lcccc}
\toprule
& \multicolumn{2}{@{}c@{}}{ZSL} & \multicolumn{2}{@{}c@{}}{GZSL} \\\cmidrule{2-3}\cmidrule{4-5}%
Method & F1 & mAP & F1 & mAP\\
\midrule
$j$ = 4 & 49.9 & 82.1 & 62.8 & 71.9\\
$j$ = 3 & 50.1 & 82.1 & 63.1 & 72.2\\
$j$ = 1 & 48.5 & 82.0 & 63.4 & 71.9\\
Ours ($j$ = 2) & \textbf{50.3} & \textbf{82.4} & \textbf{63.5} & \textbf{72.2}\\
\bottomrule
\end{tabular*}
\footnotetext{}
\end{table}
Furthermore, to analyze the effect of the number of categories generated in synthetic images on classifier training, we design experiments with the baseline method to generate multi-label synthetic images containing varying numbers of categories. Table~\ref{tb:j} demonstrates that a synthetic image containing two desired objects yields the greatest benefit to classifier training. Compared to single-label synthetic images, \textit{i.e.}, $j$ set to 1, multi-label synthetic images in our setting perform better on all evaluation metrics, especially in the ZSL task where our F1 score achieves a significant advantage. Generating more desired objects in an image, such as three or four objects, does not lead to a more significant improvement in classifier performance. Moreover, the more desired objects would increase the difficulty of the image generation framework in generating qualified images. Thus, in this paper, we set $j$ to 2, striking a balance between classifier performance and the difficulty of image generation.

\section{Conclusion}\label{sec6}
In this paper, we propose a novel image generation framework to solve the zero-shot multi-label classification challenge. We introduce a robust diffusion model as a generator to create training data for the unseen class. We incorporate a pre-trained multi-modal CLIP model as a discriminator, which serves to identify whether the generated images contain the target classes. Besides, we employ a large language model to create augmented prompts and fine-tune text encoder in diffusion model, enhancing both the diversity and efficiency of generated images. Finally, we design a global feature fusion module into the visual encoder of classification method. This module enables the visual encoder to adapt to the domain of the MLC dataset while keeping original features of the visual encoder, and helps the visual encoder to capture the global dependencies within image features. Extensive experiments are conducted on MS-COCO and NUS-WIDE datasets to validate the effectiveness of our approach in comparison to the state-of-the-art methods.

\printcredits

\section*{Declaration of competing interest}
The authors declare that they have no known competing financial interests or personal relationships that could have appeared
to influence the work reported in this paper.

\bibliographystyle{cas-model2-names}

\bibliography{cas-refs.bib}

\end{document}